\documentclass[pmlr,twocolumn,10pt]{jmlr} 

\usepackage{booktabs}
\usepackage{siunitx}

\usepackage[normalem]{ulem}
\usepackage{textcomp}

\theorembodyfont{\upshape}
\theoremheaderfont{\scshape}
\theorempostheader{:}
\theoremsep{\newline}

\newcommand{\bone}{\vec{1}}

\newcommand{\bw}{{\vec{w}}}
\newcommand{\bx}{{\vec{x}}}
\newcommand{\bX}{{\vec{X}}}
\newcommand{\by}{{\vec{y}}}
\newcommand{\bz}{{\vec{z}}}
\newcommand{\bZ}{{\vec{Z}}}
\newcommand{\bmu}{{\vec{\mu}}}
\newcommand{\bnu}{{\vec{\nu}}}
\newcommand{\bphi}{{\vec{\phi}}}

\newcommand{\bC}{{\vec{C}}}
\newcommand{\bd}{{\vec{d}}}

\newcommand{\real}{\mathbb{R}}
\newcommand{\entropy}{\mathcal{H}}

\newcommand{\norm}[1]{\bigl \| #1 \bigr\|}

\newcommand{\data}{{\set{D}}}

\newcommand{\dinternal}{\data_\mathrm{int}}
\newcommand{\dexternal}{\data_\mathrm{ext}}
\newcommand{\pint}{P_{\mathrm{int}}}  
\newcommand{\pext}{P_{\mathrm{ext}}}

\newcommand{\Zint}{\bZ}
\newcommand{\muext}{\bmu}
\newcommand{\W}{{\mathcal{W}}}
\newcommand{\wspace}[2]{\W\left( #1, #2 \right)}
\newcommand{\fdivu}[1]{D_f(#1 \| \bone/n)}
\newcommand{\klu}[1]{D_{KL}(#1 \| \bone/n)}
\newcommand{\nexternal}{n_{{\mathrm{ext}}}}
\newcommand{\ninternal}{n_{{\mathrm{int}}}}

\renewcommand{\thefootnote}{\fnsymbol{footnote}}

\DeclareMathOperator{\bernulli}{Bernoulli}
\DeclareMathOperator{\sigmoid}{sigmoid}

\title[Estimating Model Performance on External Samples]{Estimating Model Performance on External Samples from Their Limited Statistical Characteristics}

\author{%
\Name{Tal El-Hay}
\Email{talelh@kinstitute.org.il}\\
\addr KI Research Institute, Kfar Malal, Israel
\AND
\Name{Chen Yanover}
\Email{chen@kinstitute.org.il}\\
\addr KI Research Institute, Kfar Malal, Israel
}

\begin{document}

\maketitle

\begin{abstract}

Methods that address data shifts usually assume full access to multiple datasets. In the healthcare domain, however, privacy-preserving regulations as well as  commercial interests limit data availability and, as a result, researchers can typically study only a small number of datasets. In contrast, limited statistical characteristics of specific patient samples are much easier to share and may be available from previously published literature or focused collaborative efforts. 

Here, we propose a method that estimates model performance in external samples from their limited statistical characteristics. We search for weights that induce internal statistics that are similar to the external ones; and that are closest to uniform. We then use model performance on the weighted internal sample as an estimation for the external counterpart.

We evaluate the proposed algorithm on simulated data as well as electronic medical record data for two risk models, predicting complications in ulcerative colitis patients and stroke in women diagnosed with atrial fibrillation. In the vast majority of cases, the estimated external performance is much closer to the actual one than the internal performance. Our proposed method may be an important building block in training robust models and detecting potential model failures in external environments.

\end{abstract}

\paragraph*{Data and Code Availability}
This paper uses the IQVIA Medical Research Data, primary care electronic medical records (EMRs) from the United Kingdom (IMRD-UK, version: 2019-03), incorporating data from THIN, A Cegedim Database (reference made to THIN is intended to be descriptive of the data asset licensed by IQVIA), and transformed to the Observational Medical Outcomes Partnership (OMOP) common data model (CDM; v5.1) \citep{ohdsi_book_2019}. Definitions of cohorts, features, and outcomes are available through \href{https://atlas-demo.ohdsi.org/}{OHDSI demo ATLAS}. Code is available at \url{https://github.com/KI-Research-Institute/external-evaluation}.

\section{Introduction}
\label{sec:intro}
Predictive models, such as disease risk scores, are typically trained on a single, or few, data sources but are often expected to work well in other environments, that may vary in their population characteristics, clinical settings, and policies \citep{steyerberg_prediction_2016}. In many cases, model performance deteriorates significantly in these external environments, as demonstrated repeatedly (e.g., \cite{ohnuma2017prediction}), and most recently for the widely implemented proprietary Epic Sepsis Model \citep{wong_external_2021-1} and for COVID-19 risk models \citep{reps_implementation_2021-2}. 

Model robustness -- that is, its ability to provide accurate prediction despite changes, e.g., in the characteristics of input covariates -- can be demonstrated using external validation, the process of evaluating model performance on data sources that were not used for its derivation. However, full access to medical datasets is often limited due to privacy, regulatory and commercial factors. Therefore, we aim to estimate the performance of a given model on external sources using only their more commonly available statistical characteristics.  

Here, we propose an algorithm which reweights individuals in an internal sample to match external statistics, potentially reported in preceding publications or characterization studies (e.g., \citealt{recalde_characteristics_2021}); then estimates the performance on the external sample using the reweighted internal one. We focus on cases that are common in the healthcare domain, where the size of samples (that is, number of individuals) is much larger than the number of features. In such cases, infinite number of weight sets may recapitulate the external statistics, therefore the proposed algorithms searches for weights with a minimal divergence from a uniform distribution.

We first study the strengths and limitations of our suggested approach using simulated data; then split a sample from a primary care dataset into "internal" versus "external" subsets based on demographic information, and validate the approach using a prediction model for $3$-year risk of complications in ulcerative colitis patients; and, finally, use the entire primary care data to estimate the performance of three stroke risk scores in seven external resources and compare it to the actual performance, as reported in a recent study \citep{reps_feasibility_2020}. 

\section{Related Work}
\label{sec:related}

The task of evaluating model performance in external samples, often with (at least some) data shift \citep{finlayson_clinician_2021}, is tightly coupled with that of training robust models, as evaluation is a necessary step in model selection and optimization. 

One line of work handling data shifts adopts ideas from causal inference. Specifically, causal models \citep{bareinboim_causal_2016} can distinguish invariant relations between risk factors (e.g., biological or physiological) and outcomes from context- or environment-dependent mechanisms \citep{subbaswamy_preventing_2019}. \cite{subbaswamy_evaluating_2021} developed a method for analyzing model robustness (or stability) that, given a model, a set of distribution-fixed (immutable) variables and a set of distribution-varying (mutable) variables, identifies the sub-population with the worst average loss; thus, enabling evaluation of model safety, with no external information. 

Sample reweighting is commonly applied to adjust for confounders, either measured \citep{hainmueller_entropy_2012} or unmeasured \citep{streeter_adjusting_2017}, and to account for selection bias \citep{kalton_weighting_2003}, typically leveraging fully-accessible samples. Methodologically, the optimization problem we derive is similar to that studied for entropy balancing \citep{hainmueller_entropy_2012}, which attempts to reweight a sample (e.g., control group) so its prespecified set of moments exactly match that of another sample (e.g., treatment group), while maximizing the weight entropy (that is, keeping weights as close as possible to uniform). We note, however, that we explore a different use-case and, consequently, optimize over moments of an otherwise inaccessible sample (rather than samples from an accessible data source).

\section{Estimation Algorithm}
\label{sec:alg}

The goal of the proposed method is to estimate the performance of a prediction model, e.g., risk score, on an external sample, given some of its statistical properties, and using an internal, fully-accessible data. Briefly, we reweight an internal sample to obtain the external statistics, then compute model performance on the weighted sample as an estimate of the external performance.

\subsection{Problem Formulation}

Let $\bx_i$ and $y_i$ denote an observation (or feature) vector and a binary outcome\footnote[2]{We focus here on binary outcomes, as these are commonly used -- and reported -- in healthcare applications; it is possible to extend the proposed approach to continuous outcomes, using an appropriate performance measure and statistical characteristics.}, respectively, for an individual $i$. Suppose we have access to observations for $\ninternal$ individuals in an \emph{internal} sample: 
\[
\dinternal=\{ \bx_i, y_i\}_{i=1}^{\ninternal};
\]
and summary statistics for an \emph{external} sample (with $\nexternal$ individuals): 
\[  
\muext=\frac{1}{\nexternal}\sum_{(\bx_i, y_i) \in \dexternal} \bphi(\bx_i, y_i), 
\] 
where $\bphi(\bx_i, y_i)$ is a set of transformations on individual-level observations. For example: 
\[
\bphi(\bx_i, y_i) = \{ \bx_i \cdot y_i, \bx_i \cdot (1-y_i), y_i \} 
\]
allows computation of features mean in subsets of individuals with and without the outcome (as often reported in a study's Table $1$). 

We aim at estimating the performance of a model $m$ on the external sample $\dexternal$, using $\muext$ and observations from $\dinternal$. To this end, we search for weights $\bw \in [0,1]^{\ninternal}, \sum_i w_i = 1$, such that the statistical properties of the weighted sample $\left\{\bx_i, y_i, w_i \right\}_{i=1}^{\ninternal}$ approximate these of the external one. Let $\wspace{\muext}{\Zint}$ denote the space of such weight sets:
\begin{equation} \nonumber  
\begin{split}
\wspace{\muext}{\Zint} = 
\bigg\{ \bw \in \real^n : \,\, & \Zint^\top \bw = \muext, \,\, \sum w_i = 1, \,\, \\
                        & w_i \geq 0, \,\, i = 1, \ldots, \ninternal \bigg\}
\end{split}
\end{equation}
Where $\bZ$ is a matrix whose rows are $\bz_i \equiv \bphi(\bx_i, y_i)$.
As $\wspace{\muext}{\Zint}$ may be infinitely large, we propose to search for a set of weights that is also closest to uniform. 
This additional constraint is based on a \emph{proximity assumption}, intuitively that the external distribution is relatively similar to the distribution in $\wspace{\muext}{\Zint}$ that is closest to the internal distribution.

Using the reweighted sample we can now estimate two types of performance measures: 
\begin{itemize}
    \item Measures that can be expressed as a pointwise loss function, $l(m(\bx_i), y_i)$, for which we estimate the expected loss of the model in the external sample as: 
    \begin{equation} \nonumber
        \frac{1}{\ninternal}\sum_{(\bx_i, y_i) \in \dinternal} w_i \cdot l(m(x_i), y_i).
    \end{equation}
    For example, for a model that computes the probability of an outcome $y$, we can estimate the expected negative log-likelihood by setting $l(m(\bx_i), y_i) = - y_i\log(m(\bx_i)) - (1 - y_i) \log(1-m(\bx_i)).$
    
    \item Non-decomposable measures that can be evaluated on weighted samples. For example, the area under the receiver operating characteristic curve (AUC). 
\end{itemize}

Below we present a model independent scheme, which minimizes an $\emph{f-divergence}$ function (for example, maximizes the weights entropy); and in \appendixref{app:model-dep}, we derive a model (and loss) dependent scheme, which maximizes a weighted upper bound on the model loss and the regularized divergence function.

\paragraph{Model-independent optimization scheme.} 

To find a weighted representation of an internal sample that replicates the external expectations, we solve the following optimization problem:
\begin{equation} \label{eq:opt1}
\min_{\bw\in \wspace{\muext}{\Zint} }  \fdivu{\bw},
\end{equation}
where $D_f(P\|Q)$, \emph{f-divergence}, for discrete measures $P$ and $Q$ is:
\[
D_f(P\|Q) = \sum_x f\left( \frac{P(x)}{Q(x)} \right) Q(x)
\]
and $f:\real_+ \rightarrow \real$ is a convex function, with $f(1) = 0$. 
For example, when $f(t) = t \log t$, Optimization Problem (\ref{eq:opt1}) becomes: 
\begin{equation} \label{eq:opt2}
\max_{\bw\in \wspace{\muext}{\Zint}} \entropy(\bw),
\end{equation}
where $\entropy(\bw) = -\sum_i w_i \log w_i$ is the entropy function. 

We derive a dual formulation of Problem (\ref{eq:opt2}), similar to \cite{hainmueller_entropy_2012}, in \appendixref{app:dual}; and show that the optimal solution has the form:
\[
w_i \propto e^{\bz_i\cdot \bnu}\,,
\]
where $\bnu \in R^{|\phi|}$. In other words, the optimal weights are normalized exponents of a linear function of $\bZ$.
We note that, as the number of features is typically much smaller than the sample size, the solution to the dual problem is expected to be more numerically stable than the primal's.

In cases where $\wspace{\muext}{\Zint} = \varnothing$, Problem (\ref{eq:opt1}) can be adjusted to trade-off, using hyper-parameter $\lambda$, the accuracy at which the weighted internal sample reproduce the external statistics and proximity and rewritten as: 
\begin{equation} \label{eq:opt1highd}
\begin{split}
\min_{\bw} \bigg(\norm{\Zint^\top \bw - \muext} + \lambda \cdot \fdivu{\bw} \bigg) \\ 
\text{such that} \sum_i w_i = 1, w_i \geq 0\,,
\end{split}
\end{equation}
where the norm can be $L_2$ or $L_1$. 


\subsection{Detecting Estimation Failure} 
To estimate the performance of a model in an external sample $\dexternal$, with distribution $\pext(\bz)$ of transformed features, we assume that $\pint(\bz)>0$ whenever $\pext(\bz)>0$. This condition is analogous to the \emph{positivity assumption} in causal inference, except that it is \emph{one sided}. In other words, the support of $\pext(\bz)$ can be a strict subset of the support of $\pint(\bz)$. Although this assumption cannot be verified, its violations can be detected when external expectations cannot be attained in the internal sample.

\subsection{Implementation}
We used R's CVXR \citep{fu_cvxr_2020} library to solve the optimization problem and WeightedROC library \citep{hocking_weightedroc_2020} to compute weighted AUC. To deal with cases where $\wspace{\muext}{\Zint} = \varnothing$ we used the relaxed Problem (\ref{eq:opt1highd}) with $\lambda=10^{-6}$ and $L_2$ norm. To alleviate numerical issues, we set a minimum weight parameter ($10^{-6}$) and remove features with a small standard deviation ($<10^{-4}$).

\section{Empirical Evaluation}
\label{sec:emp}
To evaluate the accuracy of the proposed algorithm, we estimated the external performance of various models using an internal sample and limited external statistical characteristics, in three scenarios: (a) simulating data using a \emph{structural equation model} \citep{bareinboim_causal_2016} for "internal" and "external" environments; training an outcome prediction model on the internal sample and evaluating its performance on the external one; (b) extracting a cohort of newly diagnosed ulcerative colitis individuals in IMRD-UK data; synthetically splitting this cohort into "internal" and "external" samples; training a complication risk model on the internal sample and evaluating its performance on the external one; and (c) extracting atrial fibrillation patient cohorts in IMRD-UK data as an internal sample; evaluating the performance of three stroke risk models in multiple inaccessible claim and EMR databases using their published statistical characteristics.

\subsection{Synthetic Data}

We simulated synthetic data using structural equation models that contain a hidden variable $H\in \real$, features $\bX\in \real^p$, a binary outcome $Y$, and a deterministic binary variable $A$ where $A=0$ denotes an internal environment and $A=1$ denotes an external one (\figureref{fig:sim-model}). This framework allows examining the strengths and limitations of the proposed algorithm subject to different types of data shifts. 

\begin{figure}[htbp]
\floatconts
  {fig:sim-model}
  {\caption{Graphical representation of the data-generating causal model. $A$ is an environment variable (e.g., in a clinical setting, specific healthcare system), $H$ is a hidden variable (encoding, for example, an individual's healthcare status), $\bX$ is a set of observed features (e.g., prescribed medications or lab test results) and $Y$ is a binary outcome (e.g., disease onset or progression).}}
  {\includegraphics[width=0.5\linewidth]{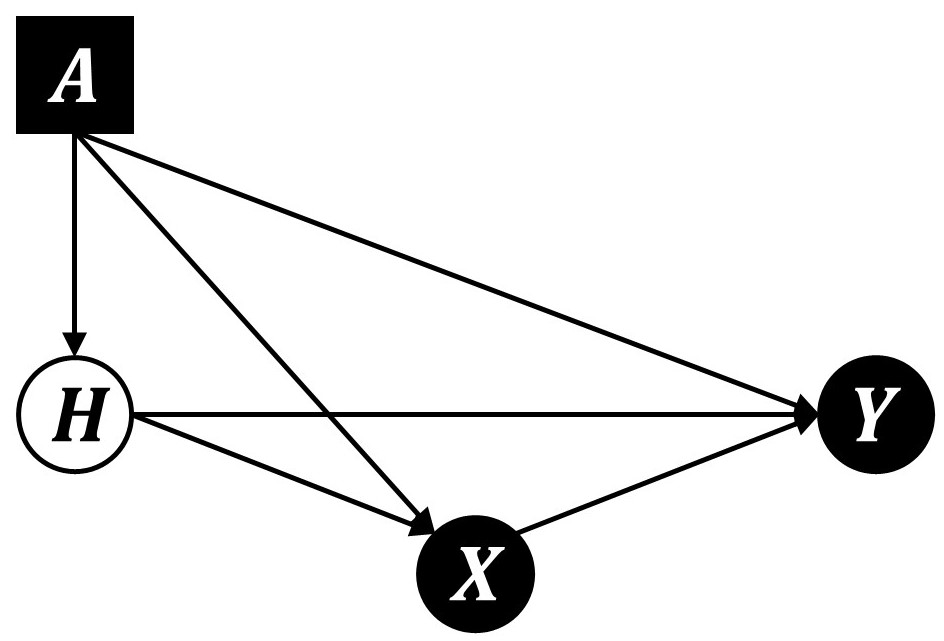}}
\end{figure} 

We defined the simulations using the following structural equations model:
\begin{equation} \nonumber
    \begin{split}
    H &= \beta_{H,A} A + \epsilon_H\\
    \bX &= \vec{\beta}_{\bX,A} A +\vec{\beta}_{\bX,H} H + \vec{\beta}_{\bX,AH} AH + \vec{\epsilon}_{\bX}\\
    Y &\sim \bernulli(\sigmoid(f(\bX, H, A)))
    \end{split}
\end{equation}
where 
\[
f(\bX, H, A) = \beta_{Y,A} A + \beta_{Y,H} H + \vec{\beta}_{Y,\vec{X}} \bX + \vec{\beta}_{Y,A\vec{X}} A\bX \,,
\]
$\beta_{\bX, \cdot}$ and $\beta_{\cdot, \bX} \in \real^p$ are coefficient vectors, the rest of the coefficients are scalars, $\epsilon_H \sim \mathcal{N}(0,1)$, ${\epsilon}_{\bX} \sim \mathcal{N}(0,\mathbf{I}_p)$ are independent sources of variability and $\sigmoid(z)=\frac{1}{1+e^{-z}}$.

This model is similar to the \emph{anchor regression model} \citep{rothenhausler_anchor_2021}, replacing the continuous outcome with a binary one. The dependency of $\bX$ on the hidden variable $H$ induces correlations between features, and the interaction term $AH$ induces differences in the correlations structure between environments. Therefore, the coefficient $\vec{\beta}_{\bX,AH}$ controls the "strength" of the shift in correlations between features, depending on the environment; and the coefficient $\vec{\beta}_{Y,A\bX}$ controls the shift in direct effect of $\bX$ on $Y$.

\subsubsection{Implementation}
Here, we set the dimension of $\bX$ to be $p=10$ and sample coefficients $\beta_{H,A}$, $\beta_{Y,A} \sim \mathcal{N}(0, 0.2)$, $\vec{\beta}_{\bX,A}\sim \mathcal{N}(0, 0.2 \mathbf{I}_p)$, $\beta_{X,H} \sim \mathcal{N}(0, \mathbf{I}_p)$, and $\beta_{Y,H} \sim \mathcal{N}(0, 1)$. We let only $X_1$ and $X_2$ (but not $X_3$ to $X_{10}$) affect the outcome $Y$ by setting $\beta_{Y,\bX} = (1, 1, 0, \ldots, 0)$ and $\beta_{Y,A\bX} = (-0.8, -0.2,  0, \ldots, 0)$.

As studies do not typically report correlations between features within each outcome class, we tested our algorithm in different scenarios of correlation shifts. Specifically, we used three configurations of $\vec{\beta}_{\bX,AH} \sim \mathcal{N}(0, \sigma_{\bX,AH})$, where $\sigma_{\bX,AH} = 0$, $0.5$, or $1$, emulating weak, medium, and strong correlation shifts, respectively.

Given a specific simulation model, we generated three data sets, namely internal training and tests sets and an external data set. We computed the mean and variance of every feature in $\bX$, separately for individuals with $Y=0$ and $Y=1$, in the external set. Next, we trained an elastic net regularized logistic regression model on the internal training set and computed the AUC on the internal test and external sets. Finally, we applied the performance estimation algorithm on the internal test set, using external expectations, and compared the estimated AUC to the actual one. 

Supplementary \figureref{fig:sim-examples} presents examples of generated samples with varying values of $\sigma_{\bX,AH}$. For each setting, we generated $200$ models, and from each sampled data with varying sizes ($n = 200$, $500$, $1000$, $2000$, $5000$). 

\subsubsection{External Performance Estimation}
The results of the proposed algorithm, in terms of divergence from uniform weights and AUC estimation accuracy, for different values of $\sigma_{\bX|AH}$ and data size $n=5000$ are shown in \tableref{tab:sim-characteristics}. As expected, weight divergence from uniform ($\klu{\bw}$) and estimation error grow with $\sigma_{X|AH}$. 

\figureref{fig:sim-error} presents the estimation error of the external AUC values, as a function of correlation shift strength and sample size, $n$. Estimation quality is lower for strong shifts in correlations which are not captured in the shared expectations, whereas milder differences result in good estimations. For comparison, the difference between internal and actual external AUC values is around $0.1$ (\tableref{tab:sim-characteristics}). 

\begin{table}[hbtp]
\floatconts
  {tab:sim-characteristics}
  {\caption{Algorithm performance in $5000$-unit simulated datasets averaged on 200 sampling repetitions. The estimation error column presents the mean of the absolute values of errors. $D_{KL}$, Kullback Leibler divergence between the derived and uniform weights, and estimation error increase with stronger correlation shifts.}}
  {\begin{tabular}{lcccc}
  \toprule
 & & Internal & External & Est. \\
  \bfseries $\sigma_{X|AH}$ & $D_{KL}$ & AUC & AUC & Error \\
  \midrule
  $0.0$ (Weak)   & $0.41$ & $0.841$ & $0.726$ & $0.011$ \\
  $0.5$ (Med.)   & $1.37$ & $0.850$ & $0.735$ & $0.019$ \\
  $1.0$ (Strong) & $4.04$ & $0.847$ & $0.733$ & $0.043$ \\
  \bottomrule
  \end{tabular}}
\end{table}

\begin{figure}[htbp]
\floatconts
  {fig:sim-error}
  {\caption{Estimation error (absolute value of the difference between actual and estimated external AUC values; y-axis) for weak, medium, and strong correlation shifts, as a function of sample size, $n$ (x-axis). Whiskers correspond to $25$-$75$ AUC percentiles, over $200$ models.}}
  {\includegraphics[width=0.8\linewidth]{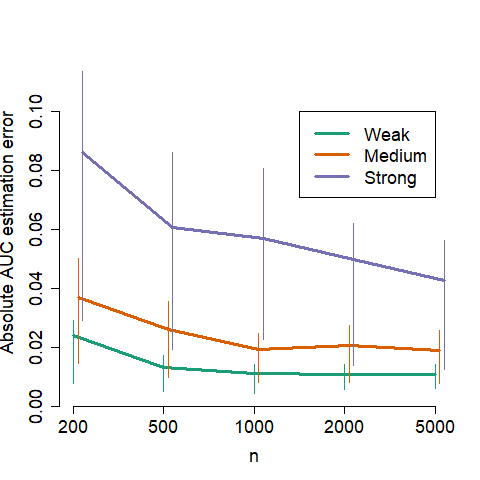}}
\end{figure} \vspace{-0.5cm}

\subsection{Synthetic Data Split: Complications of Ulcerative Colitis}
\label{subsec:UC}

Next, we studied the IMRD-UK primary care data and synthetically split it into "internal" and "external" sets based on various demographic criteria. Specifically, we trained a model on the internal sample to predict the $3$-year risk of intestinal surgery (or death) in ulcerative colitis (UC) patients; estimated its performance on the external sample, using limited external statistics; and compared the estimated and observed performance. 

\subsubsection{Clinical Background}

\begin{table*}[htbp]
\floatconts
  {tab:uc.age}%
  {\caption{Characteristics of internal and external samples, split by age.}}%
  {\begin{tabular}{lcc|cc} 
\toprule 
    & \multicolumn{2}{c|}{Internal: individuals $>34$ years old,} & 
      \multicolumn{2}{c}{Internal: individuals $\le 64$ years old,} \\ 
    & \multicolumn{2}{c|}{External: individuals $\le 34$ years old} &
      \multicolumn{2}{c}{External: individuals $>64$ years old}
       \\ \midrule
    & Internal & External & 
       Internal & External \\ \midrule 
    n & $5577$ & $1933$ & $5616$ & $1894$ \\ 
    \multicolumn{3}{l|}{Townsend deprivation index} \\
    $\,\,$ Score  & $2.4$ ($\pm 1.2$) & $2.6$ ($\pm 1.3$) & $2.5$ ($\pm 1.2$) & $2.4$ ($\pm 1.2$) \\ 
    $\,\,$ Available & $4893$ ($87.7$\%) & $1685$ ($87.2$\%) & $4913$ ($87.5$\%) & $1665$ ($87.9$\%) \\ 
    Female & $2752$ ($49.3$\%) & $932$ ($48.2$\%) & $2711$ ($48.3$\%) & $973$ ($51.4$\%) \\ 
    Smoking & $1397$ ($25$\%) & $362$ ($18.7$\%) & $1393$ ($24.8$\%) & $366$ ($19.3$\%) \\ 
    Steroids & $1597$ ($28.6$\%) & $670$ ($34.7$\%) & $1666$ ($29.7$\%) & $601$ ($31.7$\%) \\ 
    \multicolumn{3}{l|}{Body mass index (BMI)} \\
    $\,\,$ Underweight & $105$ ($1.9$\%) & $85$ ($4.4$\%) & $140$ ($2.5$\%) & $50$ ($2.6$\%) \\ 
    $\,\,$  Overweight & $1535$ ($27.5$\%) & $244$ ($12.6$\%) & $1170$ ($20.8$\%) & $609$ ($32.2$\%) \\ 
    Perianal disease & $66$ ($1.2$\%) & $49$ ($2.5$\%) & $97$ ($1.7$\%) & $18$ ($1$\%) \\ 
    Complications & $900$ ($16.1$\%) & $141$ ($7.3$\%) & $457$ ($8.1$\%) & $584$ ($30.8$\%) \\
    \bottomrule
  \end{tabular}}
\end{table*}

UC is a chronic inflammatory bowel condition with consistently increasing incidence rates in both newly industrialized and developed countries \citep{benchimol_changing_2014,windsor_evolving_2019,kaplan_four_2021}. The increase in its prevalence has a significant impact on healthcare financial burden due to chronically administered medications as well as hospitalizations and surgical procedures \citep{windsor_evolving_2019}.

UC pathogenesis is not well understood. Presumed risk factors for a more complicated disease include younger age at diagnosis, extensive disease, use of steroids and immunosupressive drugs, and being a non-smoker \citep{koliani-pace_prognosticating_2019}. 

\subsubsection{Implementation}

The UC onset cohort includes individuals with at least two diagnoses of inflammatory bowel disease (IBD) or with an IBD diagnosis and a prescription for an IBD medication; who have an ulcerative colitis diagnosis and no Crohn’s disease diagnosis. We set index (or cohort entry) date to the first IBD diagnosis or medication prescription and required that individuals have a minimum observation of $365$ days prior to index date. We excluded subjects with insufficient follow-up. 

For each individual in the ulcerative colitis cohort we extracted a set of features, previously reported as associated with increased intestinal surgery risk \citep{koliani-pace_prognosticating_2019}. These include age (and age$^2$), sex, smoking, being underweight or overweight, presence of perianal disease, and use of steroids; and considered sets of predefined features (per OHDSI's \href{https://github.com/OHDSI/FeatureExtraction}{Feature Extraction R library}), e.g., drugs prescribed to a at least 1,000 subjects up to 90 days after index date. The outcome considers procedure codes for colostomy, colectomy, ileostomy, small intestinal resection, stricturoplasty, balloon dilation, drainage of perianal abscess, drainage of intra-abdominal abscess, or death, within $3$ years following index date. 
Definition of all concept sets and cohorts are available at \url{https://atlas-demo.ohdsi.org/}.

We split the IMRD-UK data into internal and external sets based on individual age or country of living, as described below.

\begin{figure*}[tp]
\floatconts
  {fig:england-age}
  {\caption{Actual and estimated external performance in England UC cohorts, split by age. Boxes show the external median AUC and inter-quantile range (IQR, $25$ and $75$ percentiles) over $200$ repetitions; solid line represents the median internal AUC and  dashed lines represent the IQR.
  }}
  {
      \centering
      \begin{tabular}{ccccc}
          \multicolumn{2}{c}{Internal: individuals $>34$ years old,} & &  \multicolumn{2}{c}{Internal: individuals $\le 64$ years old,} \\ 
          \multicolumn{2}{c}{External: individuals $\le 34$ years old} & &  \multicolumn{2}{c}{External: individuals $>64$ years old} \\ \midrule
      {\raisebox{70pt}[0pt][0pt]{\rotatebox[origin=c]{90}{AUC}}} \,
      \subfigure[Elastic net]{\includegraphics[scale=0.21]{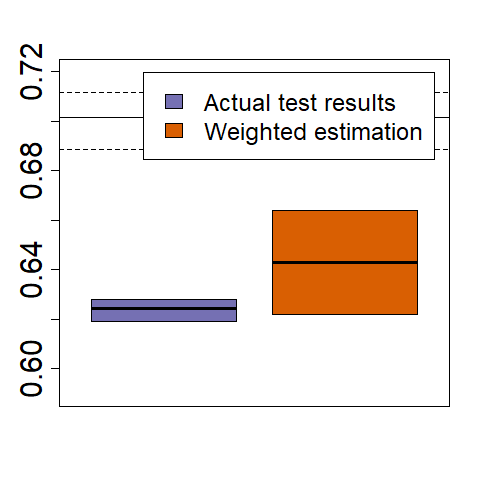}} \hspace{-.5cm} & 
      \subfigure[XGBoost]{\includegraphics[scale=0.21]{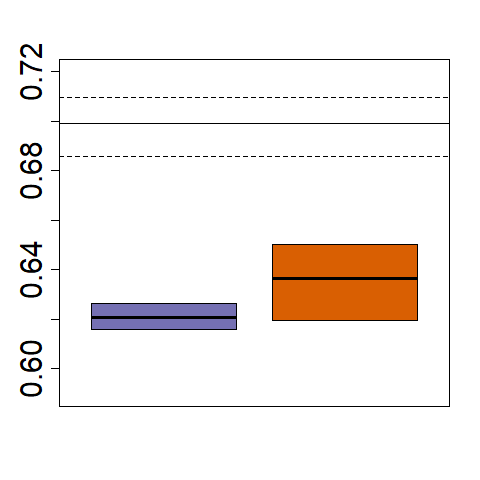}} & & 
      \subfigure[Elastic net]{\includegraphics[scale=0.21]{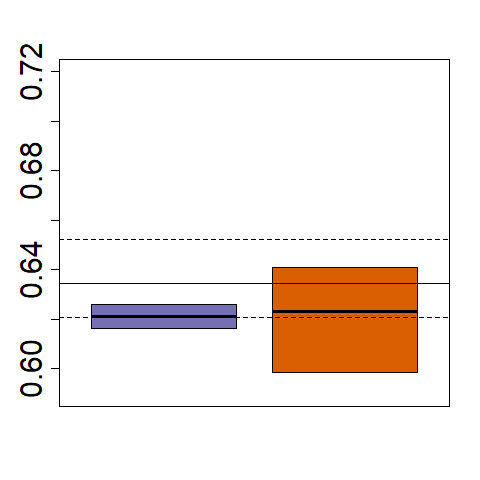}} \hspace{-.5cm} & 
      \subfigure[XGBoost]{\includegraphics[scale=0.21]{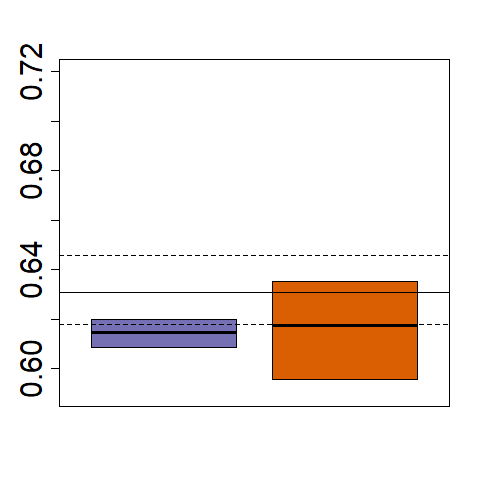}}
    \end{tabular} \\ 

  }
\end{figure*}

\subsubsection{External Performance Estimation: Ulcerative Colitis, Split by Age}

In the following experiments, we split the subset of individuals who live in England by their age. Specifically, in the first experiment, the internal set contained the $75\%$ youngest subjects ($\leq \! 64$ years) and the external set -- the $25\%$ older ones; and in the second experiment, the internal set contains the $75\%$ older individuals ($> \! 34$ years) and the internal set -- the remaining older individuals. In each of these setups we randomly split the internal set to training ($75\%$) and test ($25\%$); we repeated the training-test random split $200$ times. Next, we trained a linear model as well as a non-linear one, using XGBoost \citep{chen_xgboost_2016}, computed model's AUC on the internal test and external sets, and estimated the external AUC using the internal set and the expectations of the external one. To maintain positivity and to emulate an environment dependent hidden factor, we excluded age from the feature set. 
The populations were different in several observed characteristics, notably, percentage of women, underweight and overweight; see \tableref{tab:uc.age} for details. 

Overall, the external performance estimations, using either elastic net or XGBoost, are close to the actual ones (\figureref{fig:england-age}), notably for external younger samples, where the difference between the internal and external performance is large (right panel). 

\subsubsection{External Performance Estimation: Ulcerative Colitis, Split by Country}

\begin{table*}[htbp]
\floatconts
  {tab:uc_geog}%
  {\caption{Characteristics of ulcerative colitis, country-based sub-cohorts.}}%
  {\begin{tabular}{lcccc} \toprule 
         & England & Wales & Northern Ireland & Scotland \\ \midrule 
        n & $9469$ & $1255$ & $772$ & $1772$ \\ 
        Age (years) & $48.7$ ($\pm 18.9$) & $48.3$ ($\pm 19.1$) & $46$ ($\pm 18.2$) & $47$ ($\pm 18.5$) \\ 
        \multicolumn{5}{l}{Townsend deprivation index} \\
        $\,\,$ Score & $2.5$ ($\pm 1.2$) & $2.4$ ($\pm 1.1$) & $2.9$ ($\pm 1.3$) & $3$ ($\pm 1.2$) \\ 
        $\,\,$ Available & $8265$ ($87.3$\%) & $900$ ($71.7$\%) & $634$ ($82.1$\%) & $1541$ ($87$\%) \\ 
        Female & $4636$ ($49$\%) & $602$ ($48$\%) & $382$ ($49.5$\%) & $909$ ($51.3$\%) \\ 
        Smoking & $2230$ ($23.6$\%) & $313$ ($24.9$\%) & $221$ ($28.6$\%) & $484$ ($27.3$\%) \\ 
        Steroids & $2834$ ($29.9$\%) & $408$ ($32.5$\%) & $224$ ($29$\%) & $668$ ($37.7$\%) \\ 
        \multicolumn{4}{l}{Body mass index (BMI)} \\
        $\,\,$ Underweight & $248$ ($2.6$\%) & $30$ ($2.4$\%) & $24$ ($3.1$\%) & $37$ ($2.1$\%) \\ 
        $\,\,$ Overweight & $2276$ ($24$\%) & $343$ ($27.3$\%) & $200$ ($25.9$\%) & $442$ ($24.9$\%) \\ 
        Perianal disease & $144$ ($1.5$\%) & $16$ ($1.3$\%) & $12$ ($1.6$\%) & $11$ ($0.6$\%) \\ 
        Complications & $1315$ ($13.9$\%) & $203$ ($16.2$\%) & $123$ ($15.9$\%) & $244$ ($13.8$\%) \\ \bottomrule
  \end{tabular}}
\end{table*}

\begin{figure}[htpb]
\floatconts
    {fig:uk}
    {\caption{Actual and estimated performance in external, country-based UC samples. Boxes show the external median AUC and IQR over $200$ repetitions; solid and dashed lines represent the internal median AUC and IQR, respectively.}}
    {
      \centering
      {\small \raisebox{60pt}[0pt][0pt]{\rotatebox[origin=c]{90}{\small{AUC}}}} \, \hspace{-0.3cm}
      \subfigure[Elastic net]{\includegraphics[scale=0.235]{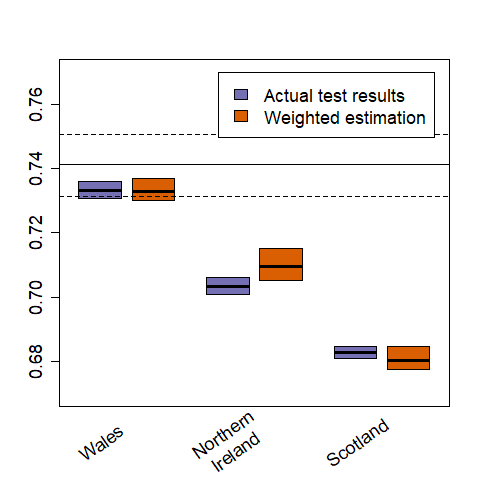}}\quad \hspace{-0.7cm}
      \subfigure[XGBoost]{\includegraphics[scale=0.235]{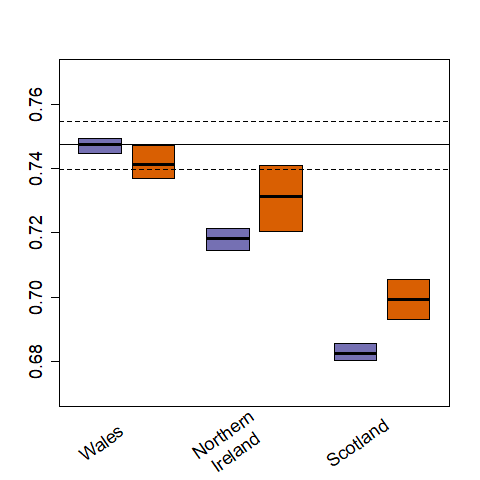}}
  }
\end{figure} 

Next, we split the UC cohort by country of residence and considered the sub-cohort of individuals living in England as the internal sample and those living in Scotland, Wales and Northern Ireland as three distinct external samples. Similarly to the age split analysis, we split the internal sample into training and test sets, repeatedly $200$ times; trained a model on each training set; extracted expectations for the external samples; and evaluated model performance on the internal test and external sets. 

The characteristics of different sub-populations are presented in \tableref{tab:uc_geog}; \figureref{fig:uk} shows the external performance evaluation results, attesting to the (much) improved accuracy of the estimated AUC values, compared to internal performance.

\subsection{Distinct Datasets: Stroke Risk Models}

\subsubsection{Clinical Background}

Atrial fibrillation is a common cardiac rhythm disorder, associated with increased risk of stroke \citep{sagris_atrial_2021}. Risk factors associated with the occurrence of stroke include older age, various comorbidities (in particular, hypertension, diabetes, and renal disease) and smoking \citep{singer_new_2013}. To guide treatment, multiple risk scores have been devised and externally evaluated in several studies \citep{van_den_ham_comparative_2015}. Recently, \cite{reps_feasibility_2020} replicated five previously published prognostic models that predict stroke in females newly diagnosed with atrial fibrillation; and externally validated their performance across nine observational healthcare datasets. Below, we use our proposed algorithm and the limited per-database statistical characteristics, as it appears in \cite{reps_feasibility_2020}, to estimate the external performance of these risk scores.

\subsubsection{Implementation}
We downloaded \cite{reps_feasibility_2020}'s \href{https://github.com/ohdsi-studies/ExistingStrokeRiskExternalValidation}{analysis package} and applied it to the IMRD-UK data, with the following modifications that adjust the study definitions to a primary care setting:
\paragraph{Target cohorts.} We considered ECG-related procedures and conditions, in addition to measurements, within 30 days prior the atrial fibrillation diagnosis, as an optional inclusion criterion.  
\paragraph{Outcome cohort.} As stroke, typically not diagnosed in a primary care setting, may be poorly recorded for deceased individuals, we added death as an entry event to the stroke cohort.
\paragraph{Feature definitions.} We extended the time window for extraction of model features to span the entire history of each individual until, and including, the date of the first atrial fibrillation event; included individuals with estimated glomerular filtration rate (eGFR) lower than $45$ mL/min/$1.73$m$^2$ in the end stage renal disease cohort, as originally defined in the ATRIA risk model \citep{singer_new_2013}; and defined former smokers as individuals with an observation of smoker, as well as those diagnosed with tobacco dependence syndrome. 

For each individual, the analysis package computed a stroke risk score given her set of features, as extracted from IMRD-UK; then, calculated score performance, vis-à-vis recorded stroke (and death) events. To estimate score performance in each external sample, we weighted individuals in the IMRD-UK data using the proposed algorithm to reproduce the sample’s populations characteristics, as reported in \cite{reps_feasibility_2020}, and computed the score performance for the weighted individual cohort. We computed $95\%$ confidence intervals using $1000$ bootstrapping iterations. 

Population attributes \citep{reps_feasibility_2020} include percentage of individuals in certain age groups ($65$-$74$ years, $75$-$85$ years and above $85$ years), comorbidities (hypertension, congestive heart failure, congestive cardiac failure, coronary heart disease, valvular heart disease, chronic and end stage renal disease, proteinuria, diabetes, and rheumatoid arthritis) and being a former smoker. 

\begin{figure*}[tp]
\floatconts
  {fig:stroke-all-ages}
  {\caption{Performance estimation for three stroke risk scores across seven external datasets. Blue circles represent the actual AUC value as reported by \cite{reps_feasibility_2020}, red diamonds show the weighted estimations, and whiskers denote $95\%$ confidence intervals. Solid lines represent the internal AUC, as computed in the IMRD-UK cohorts, with dashed lines denoting $95\%$ confidence intervals.}}
  {
      \centering
      \begin{tabular}{c} \toprule
          Women diagnosed with atrial fibrillation \\ \bottomrule
      \end{tabular} \\
      \subfigure[ATRIA]{\includegraphics[scale=0.32]{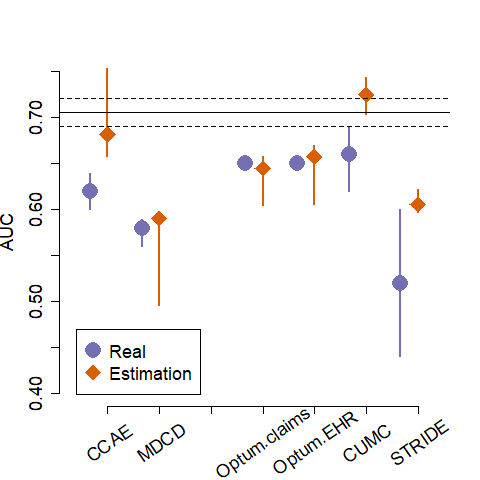}}\quad \hspace{-0.5cm}
      \subfigure[CHADS2]{\includegraphics[scale=0.32]{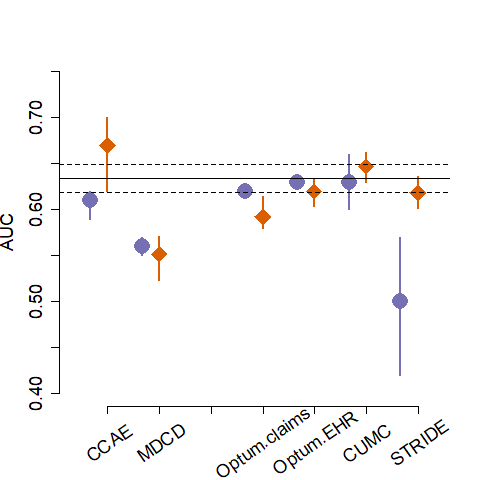}}\quad \hspace{-0.5cm}
      \subfigure[Q-Stroke]{\includegraphics[scale=0.32]{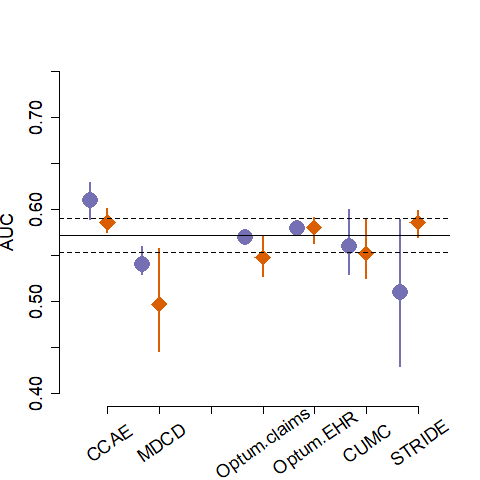}}\quad \hspace{-0.5cm} \\ 
      \begin{tabular}{c} \toprule
          Women $65$ or more years old, diagnosed with atrial fibrillation \\ \bottomrule
      \end{tabular} \\
      \subfigure[ATRIA]{\includegraphics[scale=0.32]{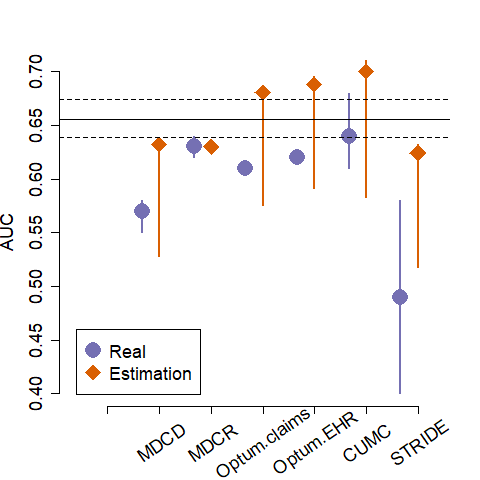}}\quad \hspace{-0.5cm}
      \subfigure[CHADS2]{\includegraphics[scale=0.32]{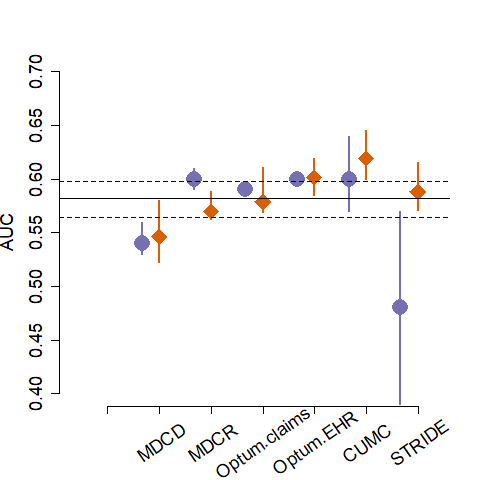}}\quad \hspace{-0.5cm}
      \subfigure[Q-Stroke]{\includegraphics[scale=0.32]{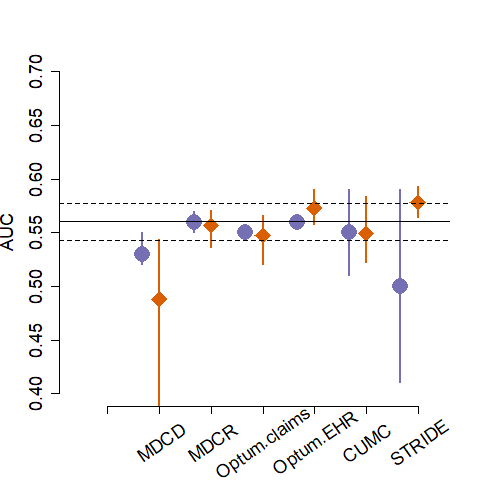}}\quad \hspace{-0.5cm}

  }
\end{figure*}

\subsubsection{External Performance Estimation}

A comparison between risk score performance, as reported by \cite{reps_feasibility_2020}, and the estimated performance is shown in \figureref{fig:stroke-all-ages}. For the full cohort (top panel), in three out of six datasets, the confidence interval of the ATRIA estimation overlaps the actual AUC (\figureref{fig:stroke-all-ages}a); in two other datasets, the estimation is better than the internal, IMRD-UK based performance. Qualitatively similar results are observed for the CHADS2 and Q-Stroke risk scores (\figureref{fig:stroke-all-ages}b and c, respectively); as well as for women $65$ years or older (bottom panel).

We note that for two additional risk scores, Framingham and CHA$_2$DS$_2$VASc, and two datasets, Ajou University School Of Medicine (AUSOM) and Integrated Primary Care Information (IPCI), \cite{reps_feasibility_2020} do not provide necessary statistical characteristics. Additionally, AUC values of IBM MarketScan\textsuperscript{\textregistered} Medicare Supplemental Database (MDCR) were not provided for the full atrial fibrillation cohort and those of IBM MarketScan\textsuperscript{\textregistered} Commercial Database (CCAE) are missing for the older female sub-cohort. Therefore, in all those cases, performance estimations are not reported. 

\section{Discussion}
We presented an algorithm that estimates the performance of prediction models on external samples from their limited statistical characteristics; and demonstrated its utility using synthetic data, synthetic split of an ulcerative colitis cohort from a single database into age groups and by country of living, and a recent risk model benchmark of stroke risk models on multiple external samples. Importantly, our proposed algorithm can help identifying models that perform well across multiple clinical settings and geographies, even when detailed test data from such settings is not available. It can thus direct development of robust models and accelerate deployment to external environments. 

The algorithm relies on two assumptions: one-sided positivity and proximity. Both assumptions cannot be fully tested, but clear violations of the former one can be detected, for example, when the expected value of a feature is non-zero in the external distribution but all the individuals in the internal set have a zero value for that feature. Intuitively, proximity is more likely to be plausible when the statistical information becomes more detailed. Therefore, whereas our preliminary experiments involved only marginal statistics of features it may be informative to test the performance of the algorithm when more detailed statistics are available, for example interactions among features or information available in deep characterization studies \cite{burn_deep_2020-1}.

We believe that the proposed methodology can serve as a building block in network studies that aim to construct robust models across datasets when data sharing is limited, e.g., by regulatory constraints. Although federated learning methods may be a promising avenue for such scenarios, it would be interesting to explore in which cases the proposed algorithm can facilitate a one-shot federated learning scheme, that does not require deployment of federated algorithm clients in all network nodes. 

In future work, we will combine the proposed algorithm with methods that aim to construct robust models such as those that leverage distributionally robust optimization 
\citep{buhlmann_invariance_2020};
study methods that exploit the relations between calibration and robustness \citep{wald_calibration_2022}; and look into decomposing AUC \citep{pmlr-v54-eban17a}, so it can be optimized explicitly.

\section*{Institutional Review Board (IRB)}
This study has been approved by IQVIA Scientific Review Committee (Reference numbers: 21SRC066, 22SRC002).

\acks{We thank Drs Roni Weisshof and Ramit Magen, Rambam Health Care Campus, Haifa, Israel, for their help in defining the ulcerative colitis predictive model; and Prof Seng Chan You, Yonsei University Health System, Seoul, Republic of Korea, for pointing us to the stroke external validation study. We are also grateful to KI Institute's researchers and, specifically, to Nir Kalkstein and Yonatan Bilu, for many fruitful discussions.}

\bibliography{robustness}

\appendix

\section{Model-dependent optimization scheme}
\label{app:model-dep}
An upper bound of a model $m$'s weighted loss $l$, up to a finite sample error, can be derived as follows: 
\[
\max_{\bw\in\wspace{\muext}{\Zint}} \sum_i w_i \cdot l(m(x_i), y_i).
\]

The tightness of the bound may depend on the number of expectations we consider. Furthermore, as $\bz$, and consequently $\muext$, may not represent all inter-feature dependencies existing in the data, an additional constraint may yield improved estimations:
\begin{equation}
\label{eqn:reg-max-objective}
\max_{\bw\in\wspace{\muext}{\Zint}} \sum_i w_i \cdot l(m(x_i), y_i) - \lambda\fdivu{\bw}. 
\end{equation}
As we increase $\lambda$, the bound may become tighter but confidence may decrease.

\section{Model-independent dual optimization problem}
\label{app:dual}
Recall that optimization Problem (\ref{eq:opt2}) is defined as follows: 
\begin{equation}
\begin{split}
\textrm{minimize}_w \quad & -\entropy(\bw) \, \,\, \\
\textrm{such that}\quad   & \bZ^\top \bw = \bmu, \bone^\top \bw = 1
\end{split}
\end{equation}
where $\bw \geq 0$. Denoting
\begin{equation}
\bC  = \left[ 
	\begin{array}{c}
	\bZ^\top \\
	\bone^\top
	\end{array}
	\right], \quad
\bd = \left[ 
	\begin{array}{c}
	\bmu \\
	1
	\end{array}
	\right],
\end{equation}
Problem (\ref{eq:opt2}) becomes: 
\begin{equation}
\begin{split}
\textrm{minimize}_w \quad	& -\entropy(\bw) \, \,\, \\
\textrm{such that}  \quad	& \bC^\top \bw = \bd
\end{split}
\end{equation}

Following Equation 5.11 in \cite{boyd2004convex} the dual function is:
\[
g(\bnu) = -\bd^\top \bnu  - (- \entropy)^*(-\bC^\top \bnu)
\]
where $(-\entropy)^*$ is the conjugate of the negative-entropy function (\cite{boyd2004convex}, p. 222):
\[
(- \entropy)^*(\by) = \sum_{i=1}^n e^{y_i - 1}
\]
Therefore,
\[
g(\bnu) = -(\bmu, 1)^\top \bnu  - e^{-1} \sum_{i=1}^n e^{-(\bz_i, 1) \bnu}
\]

The Lagrangian of the primal problem is: 
\begin{equation}
L(\bw; \bnu) = \sum_i w_i \log w_i + \bnu^\top(\bC \bw - \bd)
\end{equation}
Let $\bnu^*$ be the optimal solution of $\max_\bnu g(\bnu)$. Then, following Section 5.5.3 of \cite{boyd2004convex}, the solution of the primal problem minimizes the Lagrangian at $\bnu^*$:
\[
\frac{\partial L(\bw; \bnu^*)}{\partial w_i} = \log w_i + 1 + (\bz_i, 1) \; \bnu^* = 0
\]
giving
\[
w_i = e^{-1-(\bz_i, 1) \bnu^*} \,.
\]
This result shows that the optimal weights are normalized exponents of a linear function of the data points.

\begin{figure*}[htpb]
\section{Supplementary Figures}
\floatconts
  {fig:sim-examples}
  {\caption{Simulation examples with varying values of $\sigma_{X,AH}$. Dot colors denote outcome class, diamonds represent class means. The shift in correlation between $X_1$ and $X_2$, given an outcome class, increases with $\sigma_{X,AH}$.}}
  {
      \centering
      \subfigure[$\sigma_{X,AH}=0$ internal]{\includegraphics[scale=0.35]{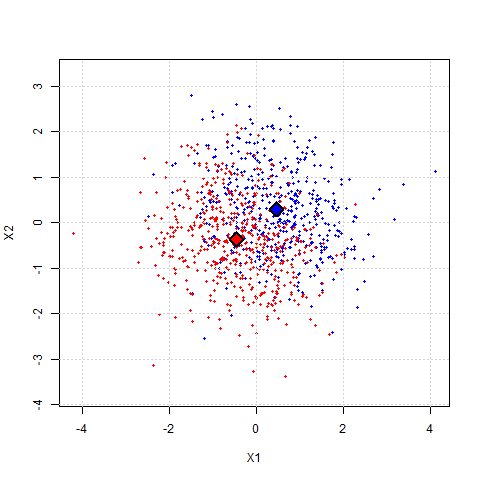}}\quad
      \subfigure[$\sigma_{X,AH}=0$ external]{\includegraphics[scale=0.35]{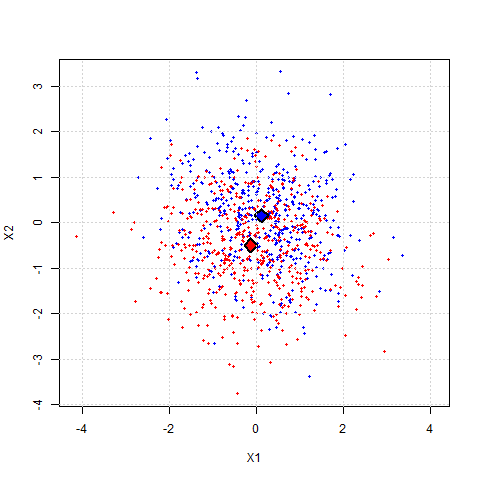}}\quad \\ 
      \subfigure[$\sigma_{X,AH}=0.5$ internal]{\includegraphics[scale=0.35]{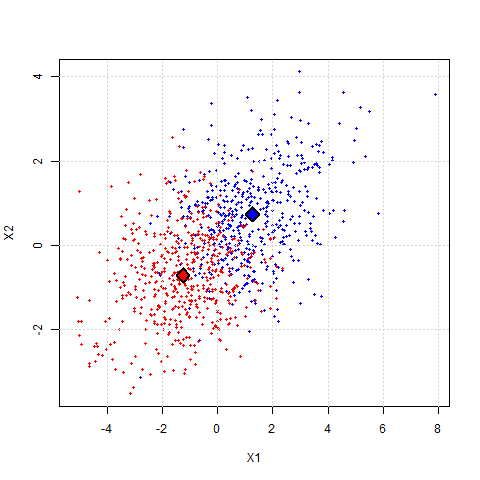}}\quad
      \subfigure[$\sigma_{X,AH}=0.5$ external]{\includegraphics[scale=0.35]{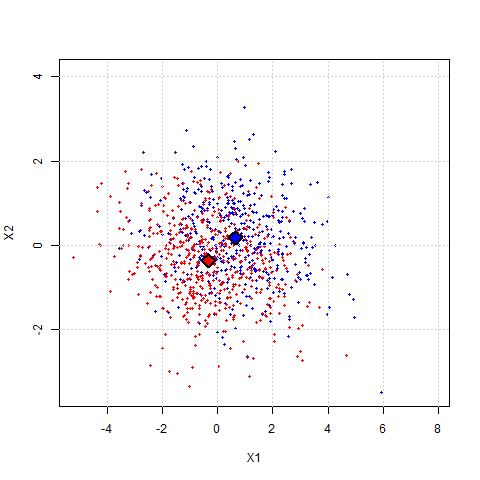}} \\ 
      \subfigure[$\sigma_{X,AH}=1$ internal]{\includegraphics[scale=0.35]{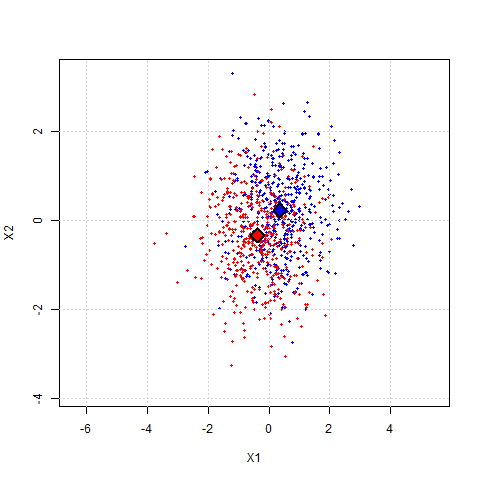}}\quad
      \subfigure[$\sigma_{X,AH}=1$ external]{\includegraphics[scale=0.35]{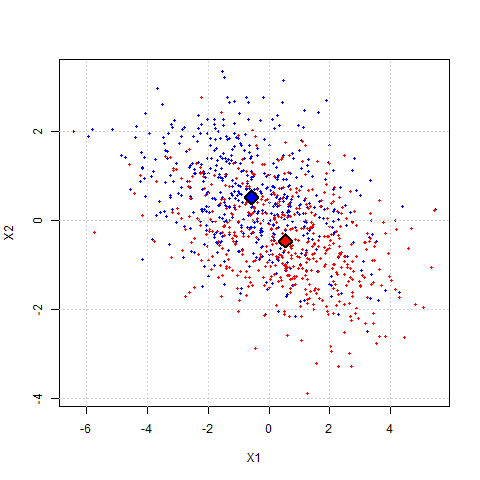}}
  }
\end{figure*}

\end{document}